\documentclass[nonacm,sigplan]{acmart}

\usepackage{url}
\usepackage{hyperref}
\usepackage[hyphenbreaks]{breakurl}

\usepackage{multirow}
\usepackage{bbm}
\usepackage{verbatimbox}
\usepackage[frozencache,cachedir=.]{minted}
 
\usepackage{amsmath}

\usepackage[linesnumbered,ruled,vlined]{algorithm2e}
\SetKwInput{KwInput}{Input}     
\SetKwInput{KwOutput}{Output}    
\SetKwInput{KwParameter}{Parameter}
\SetKw{KwBreak}{break}  

\usepackage{tcolorbox}
\tcbuselibrary{minted,breakable,xparse,skins}

\AtBeginDocument{%
  \providecommand\BibTeX{{%
    \normalfont B\kern-0.5em{\scshape i\kern-0.25em b}\kern-0.8em\TeX}}}

\begin{document}

\title{biquality-learn: a Python library for Biquality Learning}
\author{Pierre Nodet}
\email{pierre.nodet@orange.com}
\affiliation{
  \institution{Orange Innovation, Université Paris-Saclay}
  \city{Châtillon}
  \country{France}
  \postcode{92320}
}

\author{Vincent Lemaire}
\email{vincent.lemaire@orange.com}
\affiliation{
  \institution{Orange Innovation}
  \city{Lannion}
  \country{France}
  \postcode{22300}
}

\author{Alexis Bondu}
\email{alexis.bondu@orange.com}
\affiliation{
  \institution{Orange Innovation}
  \city{Châtillon}
  \country{France}
  \postcode{92320}
}

\author{Antoine Cornuéjols}
\email{antoine.cornuejols@agroparistech.fr}
\affiliation{
  \institution{AgroParisTech, INRAe, Université Paris-Saclay}
  \city{Palaiseau}
  \country{France}
  \postcode{91120}
}

\renewcommand{\shortauthors}{Nodet, et al.}

\begin{abstract}
The democratization of Data Mining has been widely successful thanks in part to powerful and easy-to-use Machine Learning libraries.  These libraries have been particularly tailored to tackle Supervised Learning. However, strong supervision signals are scarce in practice, and practitioners must resort to weak supervision. In addition to weaknesses of supervision, dataset shifts are another kind of phenomenon that occurs when deploying machine learning models in the real world.  That is why Biquality Learning has been proposed as a machine learning framework to design algorithms capable of handling multiple weaknesses of supervision and dataset shifts without assumptions on their nature and level by relying on the availability of a small trusted dataset composed of cleanly labeled and representative samples. Thus we propose biquality-learn: a Python library for Biquality Learning with an intuitive and consistent API to learn machine learning models from biquality data, with well-proven algorithms, accessible and easy to use for everyone, and enabling researchers to experiment in a reproducible way on biquality data.

\end{abstract}

\keywords{Python, Biquality Learning, Weakly Supervised Learning, Dataset Shift}

\maketitle

\section{Introduction}

The democratization of Data Mining has been widely successful thanks in part to powerful and easy to use Machine Learning libraries such as scikit-learn \cite{scikit-learn}, weka \cite{witten2002data}, or caret \cite{JSSv028i05}. These libraries have been at the core of enforcing good practices in Machine Learning and providing efficient solutions to complex problems. These libraries have been particularly tailored to tackle Supervised Learning and occasionally Semi-Supervised Learning and Unsupervised Learning. However, strong supervision signals are scarce in practice, and practitioners must resort to weak supervision. Learning with weak supervisions, or Weakly-Supervised Learning \cite{zhou2017}, is a diverse field, as diverse as the identified weaknesses of supervision. Usually, weaknesses of supervision are divided into three groups, namely \textit{inaccurate supervision} when samples are mislabeled, \textit{inexact supervision} when labels are not adapted to the classification task, or \textit{incomplete supervision} when labels are missing which reflects the inadequacy of the available labels in the real world \cite{JSSv028i05}. For each weakness of supervision, algorithms have to be specifically hand designed to alleviate them. In addition to weaknesses of supervision, dataset shifts are another kind of phenomenon that occurs when deploying machine learning models in the real world \cite{quinonero2008dataset}. Dataset shifts happen when the data distribution observed at training time differs from what is expected from the data distribution at testing time \citep{moreno2012unifying}. Shifts in the joint distribution of features and targets can be further divided into four subgroups of shifts, \textit{covariate shift} for shifts in the feature distribution, \textit{prior shift} for shifts in the target distribution, \textit{concept drift} for shifts in the decision boundary, and \textit{class-conditional shift} for shifts in the feature distribution for a given target. Again, designing algorithms to handle dataset shifts usually requires assumptions on the nature of the shift \citep{david2010impossibility}. Because of the diverse nature of possible weaknesses of supervision and dataset shifts, and robust algorithms' associated assumptions, it is impossible for practitioners to choose the suited approach to their problem.

Biquality Learning is a machine learning framework that has been proposed to design algorithms capable of handling multiple weaknesses of supervision, and dataset shifts without assumptions on their nature \cite{nodet2021weakly}. It relies on the availability of a small \textit{trusted dataset} composed of cleanly labeled and representative samples for the targeted classification task, in addition to the usual \textit{untrusted dataset} composed of potentially corrupted and biased samples. Even though the trusted dataset is not big or rich enough to properly learn the targeted classification task, it is sufficient to learn a mapping function from the untrusted distribution to the trusted distribution to train machine learning models on corrected untrusted samples.

Leveraging trusted data has proven to be particularly efficient to combat distribution shifts \cite{Hendrycks2018,fang2020rethinking} especially on the most engaging corruptions such as instance-dependant label noise \cite{nodet2021importance}. In many real-world scenarios, these trusted data are available or can easily be made available to use Biquality Learning algorithms to train robust machine learning models. One occurrence is when annotating an entire dataset is expansive to the point of being prohibitive, but labeling a small part of the dataset is doable. In Fraud Detection and Cyber Security, labeling samples require complex forensics from domain experts, limiting the number of clean samples. However, the rest of the dataset can be labeled by hand-engineered rules \citep{ratner2020snorkel} with labels that cannot properly be trusted. Another scenario happens where data shifts happen during the labeling process over time. It arises in MLOps \citep{kreuzberger2022machine}, when a model is first learned on clean data and then deployed in production, or when past predictions are used to learn an updated model \citep{7502263}. Finally, when multiple annotators are responsible for dataset labeling, which happens in NLP, the annotators' efficiency in following these guidelines may vary. However, suppose one annotator can be trusted. In that case, all the other annotators can be considered untrusted, and associating each untrusted annotator against the trusted annotator can be viewed as a Biquality Learning task \citep{yuen2011survey}.

Multiple libraries have been developed recently for the purpose of handling covariate shift, especially for Domain Adaptation \cite{de2021adapt} or for dealing with weak supervisions \cite{10.1007/978-3-031-21244-4_5}. However, Biquality Learning lacks an accessible library with an intuitive and consistent API to learn machine learning models from Biquality Data, with well-proven algorithms. Thus we propose \textbf{biquality-learn}: a Python library for Biquality Learning.

\section{biquality-learn}

We designed the \textbf{biquality-learn} library following the general design principles of \textbf{scikit-learn}, meaning that it provides a consistent interface for training and using biquality learning algorithms with an easy way to compose building blocks provided by the library with other blocks from libraries sharing these design principles \cite{buitinck2013api}. It includes various reweighting algorithms, plugin correctors, and functions for simulating label noise and generating sample data to benchmark biquality learning algorithms.

\textbf{biquality-learn} and its dependencies can be easily installed through pip:
\begin{verbatim}
pip install biquality-learn
\end{verbatim}

Overall, the goal of \textbf{biquality-learn} is to make well-known and proven biquality learning algorithms accessible and easy to use for everyone and to enable researchers to experiment in a reproducible way on biquality data.

\begin{itemize}
    \item Source Code: \url{https://github.com/biquality-learn/biquality-learn}
    \item Documentation: \url{https://biquality-learn.readthedocs.io/}
    \item License: BSD 3-Clause
\end{itemize}

\section{Design of the API}
\label{design}

Scikit-learn \cite{scikit-learn} is a machine learning library for Python with a design philosophy emphasizing consistency, simplicity, and performance. The library provides a consistent interface for various algorithms, making it easy for users to switch between models. It also aims to make machine learning easy to get started with through user-friendly API and precise documentation. Additionally, it is built on top of efficient numerical libraries (numpy \cite{2020NumPy-Array}, and SciPy \cite{2020SciPy-NMeth}) to ensure that models can be trained and used on large datasets in a reasonable amount of time. 

In \textbf{biquality-learn}, we followed the same principle, implementing a similar API with \textit{fit}, \textit{transform}, and \textit{predict} methods. In addition to passing the input features $X$ and the labels $Y$ as in \textbf{scikit-learn}, in \textbf{biquality-learn}, we need to provide information regarding whether each sample comes from the trusted or trusted or untrusted dataset. We require an additional \textit{sample\_quality} untrusted dataset: the additional \textit{sample\_quality} parameter serves to specify property to specify from which dataset the sample originates. Especially from which dataset the sample originates where a value of 0 indicates an untrusted a value of 0 indicates an untrusted sample, and 1 indicates a trusted sample, and 1 a trusted one.


\section{Algorithms Implemented}

In \textbf{biquality-learn}, we purposely implemented only a specific class of algorithms centered on approaches for tabular data and classifiers, thus restricting approaches that are genuinely classifier agnostic or implementable within \textbf{scikit-learn}'s API. We did so not to break the design principles shared with \textbf{scikit-learn} and not impose a particular deep learning library such as PyTorch \cite{NEURIPS2019_9015}, or TensorFlow \cite{tensorflow2015-whitepaper} on the user.

We summarized all implemented algorithms and what kind of corruption they can handle in the following Table.

\begin{table}[!h]
\centering
\resizebox{\columnwidth}{!}{
\begin{tabular}{@{}lcc@{}}
\toprule
\multirow{2}{*}{\textbf{Algorithms}}     & \textbf{Dataset} & \textbf{Weaknesses} \\
     & \textbf{Shifts} & \textbf{of Supervision} \\
\midrule
EasyAdapt \cite{daume2009frustratingly} &  $\checkmark$ &  $\times$\\
TrAdaBoost \cite{dai2007boosting} &  $\checkmark$ &  $\times$\\
Unhinged (Linear/Kernel) \cite{NIPS2015_5941}&  $\times$  & $\checkmark$ \\
Backward \cite{natarajan2017cost,patrini2017making} (with GLC \cite{Hendrycks2018}) & $\times$ & $\checkmark$  \\
IRLNL \cite{liu2015classification, wang2017multiclass} (with GLC \cite{Hendrycks2018}) & $\times$ & $\checkmark$  \\
Plugin \cite{zhang2021learning} (with GLC \cite{Hendrycks2018}) & $\times$ & $\checkmark$  \\
$K$-KMM \cite{fang2020rethinking} & $\checkmark$ & $\checkmark$\\
IRBL \cite{nodet2021importance} & $\times$ & $\checkmark$\\
$K$-PDR & $\checkmark$ & $\checkmark$\\
 \bottomrule
\end{tabular}
}
\caption{Algorithms Implemented in biquality-learn}
\label{table-bqlearn}
\end{table}

\section{Training Biquality Classifiers}

Training a biquality learning algorithm using \textbf{biquality-learn} is the same procedure as training a supervised algorithm with \textbf{scikit-learn} thanks to the design presented in Section \ref{design}. The features $X$ and the targets $Y$ of samples belonging to the trusted dataset $D_T$ and untrusted dataset $D_U$ must be provided as one global dataset $D$. Additionally, the indicator representing if a sample is trusted or not has to be provided: $\textit{sample\_quality}=\mathbbm{1}_{X\in D_T}$. 

Here is an example of how to train a biquality classifier using the $K$-KMM ($K$-Kernel Mean Matching) \cite{fang2020rethinking} algorithm from \textbf{biquality-learn}:\\

\begin{verbnobox}[\fontsize{9pt}{9pt}\selectfont]
from sklearn.linear_models import LogisticRegression
from bqlearn.kdr import KKMM

kkmm = KKMM(kernel="rbf", LogisticRegression())
kkmm.fit(X, y, sample_quality=sample_quality)
kkmm.predict(X_new)   
\end{verbnobox}

\section{scikit-learn's metadata routing}

\textbf{scikit-learn}'s metadata routing is a Scikit Learn Enhancement Proposal (SLEP006) describing a system that can be used to seamlessly incorporate various metadata in addition to the required features and targets in \textbf{scikit-learn} estimators, scorers and transformers. \textbf{biquality-learn} uses this design to integrate the \textit{sample\_quality} property into the training and prediction process of biquality learning algorithms. It allows one to use \textbf{biquality-learn}'s algorithms in a similar way to \textbf{scikit-learn}'s algorithms by passing the \textit{sample\_quality} property as an additional argument to the \textit{fit}, \textit{predict}, and other methods.

Currently, the main components provided by \textbf{scikit-learn} support this design and is already usable for cross-validators. However, it will be extended to all components in the future, and \textbf{biquality-learn} will significantly benefit from many ``free'' features. When \url{https://github.com/scikit-learn/scikit-learn/pull/24250} will be mer\-ged, it will be possible to make a bagging ensemble of biquality classifiers thanks to the \textit{BaggingClassifier} implemented in \textbf{scikit-learn} without overriding its behavior on biquality data.\\

\begin{verbnobox}[\fontsize{9pt}{9pt}\selectfont]

from sklearn.ensemble import BaggingClassifier

bag = BaggingClassifier(kkmm).fit(X, y, 
                    sample_quality=sample_quality)
\end{verbnobox}

\section{Cross-Validating Biquality Classifiers}

Any cross-validators working for usual Supervised Learning can work in the case of Biquality Learning. However, when splitting the data into a train and test set, untrusted samples need to be removed from the test set to avoid computing supervised metrics on corrupted labels. That is why \textit{make\_biquality\_cv} is provided by \textbf{biquality-learn} to post-process any \textbf{scikit-learn} compatible cross-validators.

Here is an example of how to use \textbf{scikit-learn'}s \textit{RandomizedSearchCV} function to perform hyperparameter validation for a biquality learning algorithm in \textbf{biquality-learn}:\\

\begin{verbnobox}[\fontsize{8pt}{8pt}\selectfont]

from sklearn.model_selection import RandomizedSearchCV
from sklearn.utils.fixes import loguniform
from bqlearn.model_selection import make_biquality_cv

param_dist = {"final_estimator__C": loguniform(1e3, 1e5)}
n_iter=20

random_search = RandomizedSearchCV(
   kkmm,
   param_distributions=param_dist,
   n_iter=n_iter,
   cv=make_biquality_cv(X, sample_quality, cv=3)
)
random_search.fit(X, y, sample_quality=sample_quality)
\end{verbnobox}

\section{Simulating Corruptions with the Corruption API}

The \textit{corruption} module in \textbf{biquality-learn} provides several functions to artificially create biquality datasets by introducing synthetic corruption. These functions can be used to simulate various types of label noise or imbalances in the dataset. We hope to ease the benchmark of biquality learning algorithms thanks to the corruption API, with a special touch on the reproducibility and standardization of these benchmarks for researchers.

Here is a brief overview of the functions available in the corruption module:
\begin{itemize}
    \item \textit{make\_weak\_labels}: Adds weak labels to a dataset by learning a classifier on a subset of the dataset and using its predictions as a new label.
   \item \textit{make\_label\_noise}: Adds noisy labels to a dataset by randomly corrupting a specified fraction of the samples thanks to a given noise matrix.
   \item \textit{make\_instance\_dependent\_label\_noise}: Adds instance-de\-pen\-dent noisy labels by corrupting samples with a probability depending on the sample and a given noise matrix.
   \item \textit{uncertainty\_noise\_probability}: Computes the probability of corrupting a sample based on the prediction uncertainty of a given classifier \cite{nodet2021importance}.
   \item \textit{make\_feature\_dependent\_label\_noise}: Adds instance-de\-pen\-dent noisy labels by corrupting a specified fraction of the labels with a probability depending on a random linear map between the features space and the labels space \cite{xia2020part}.
   \item \textit{make\_imbalance}: Creates an imbalanced dataset by oversampling or undersampling the minority classes \cite{buda2018systematic}.
   \item \textit{make\_sampling\_biais}: Creates a sampling biais by sampling not at random a subset of the dataset from the original dataset. The sampling scheme follows a Gaussian distribution with a shifted mean and scaled variance computed from the first principal component of a PCA learned from the dataset \cite{gretton2009covariate}.
\end{itemize}

\section{Conclusion}

We presented \textbf{biquality-learn}, a Python library for Biquality Learning. We exposed the design behind its API to make it easy to use and consistent with \textbf{scikit-learn}. We notably showed the future-proof of our design by showing how well it integrates with the future design of \textbf{scikit-learn}. In the future, \textbf{biquality-learn} could be supported with deep learning capabilities with a twin library with a principled design, committing to a deep learning library. Finally, the capacity of \textbf{biquality-learn} could be extended to particularly needed capabilities in real-world scenarios, such as evaluating machine learning models on untrusted data.


\bibliographystyle{ACM-Reference-Format}
\bibliography{references.bib}


\end{document}